\documentclass{article}
\usepackage{ijcai}

\usepackage{times}
\usepackage{xcolor}
\usepackage{soul}
\usepackage{url}
\usepackage[hidelinks]{hyperref}
\usepackage[utf8]{inputenc}
\usepackage[small]{caption}
\usepackage{graphicx}
\usepackage{amsmath}
\usepackage{amsthm}
\usepackage{booktabs}
\usepackage{algorithm}
\usepackage{algcompatible}
\usepackage{subcaption}
\graphicspath{ {./images/} }

\usepackage{titlesec}
\usepackage{graphicx}
\usepackage{amsmath}
\newcommand{\norm}[1]{\left\lVert#1\right\rVert}
\usepackage{amsmath,amssymb,amsfonts}
\usepackage{array}
\usepackage{color}
\usepackage{enumitem}
\usepackage{float}
\usepackage{svg}

\usepackage[toc,page]{appendix}

\newcommand{\twopartdef}[4]
{
	\left\{
	\begin{array}{ll}
		#1 & #2 \\
		#3 & #4
	\end{array}
	\right.
}

\newcommand{\specialcell}[2][c]{%
  \begin{tabular}[#1]{@{}c@{}}#2\end{tabular}}

\titleformat{\paragraph}
{\normalfont\normalsize\bfseries}{\theparagraph}{1em}{}
\titlespacing*{\paragraph}
{0pt}{3.25ex plus 1ex minus .2ex}{1.5ex plus .2ex}

\newtheorem{example}{Example}

\newtheorem{definition}{Definition}
\newtheorem{protocol}{Protocol}
\newtheorem{remark}{Remark}
\newtheorem{property}{Property}
\newtheorem{observation}{Observation}
\newtheorem{problem}{Problem}

\title{FedXGBoost: Privacy-Preserving XGBoost for Federated Learning}

\author{
Nhan Khanh Le$^1$\footnote{Contact Author}\and
Yang Liu$^2$\and 
Quang Minh Nguyen$^4$\and
Qingchen Liu$^{1}$\and
Fangzhou Liu$^{3}$\and
Quanwei Cai$^2$
\And
Sandra Hirche$^{1}$
\affiliations
$^1$Chair of Information-Oriented Control, Technical University of Munich\\
$^2$Security Research, Bytedance Inc.\\
$^3$Chair of Automatic Control Engineering, Technical University of Munich\\
$^4$ Department of EECS, Massachusetts Institute of Technology 
\emails
liuyang.fromthu@bytedance.com, caiquanwei@bytedance.com,
\{nhankhanh.le, qingchen.liu, fangzhou.liu, hirche\}@tum.de, nmquang@mit.edu
}

\begin{document}
\maketitle
\begin{abstract}
Federated learning is the distributed machine learning framework that enables collaborative training across multiple parties while ensuring data privacy. Practical adaptation of XGBoost, the state-of-the-art tree boosting framework, to federated learning remains limited due to high cost incurred by conventional privacy-preserving methods. To address the problem, we propose two variants of federated XGBoost with privacy guarantee: \emph{FedXGBoost-SMM} and \emph{FedXGBoost-LDP}. Our first protocol \emph{FedXGBoost-SMM} deploys enhanced secure matrix multiplication method to preserve privacy with lossless accuracy and lower overhead than encryption-based techniques. Developed independently, the second protocol \emph{FedXGBoost-LDP} is heuristically designed with noise perturbation for local differential privacy, and empirically evaluated on real-world and synthetic datasets. 
\end{abstract}

\section{Introduction}

As a distributed machine learning (ML) model, FL  benefits from the variety of multiple data holders and  facilitates collaborative training. Its widespread usage can be found in credit card fraud detection \cite{WensiFL}, banking prediction \cite{GeetFL}, and health care application by \cite{XuFL}.
Nevertheless, participants of a FL model are required to share the knowledge of their data, leading to the threat of privacy leakage. \emph{Privacy-preserving} in \emph{Federated Learning } (FL) is thus one of the key challenges. 
Several anonymization approaches that cover users' identification are shown to be  insufficient  \cite{Narayanan2006HowTB}. Furthermore, the European Union recently imposed General Data Protection Regulation (GDPR) to increase the privacy protection of user's private data. Therefore, any FL framework must satisfy privacy-preserving criteria while offering high-quality ML service. An informative overview of privacy-preserving under FL is provided in \cite{Qiang19}. 

XGBoost - Gradient boosted decision trees by \cite{Chen} is a novel tree ensemble model that achieves state-of-the-art results on a variety of machine learning problems. In this paper, we aim to bring the benefits of XGBoost to the FL settings with privacy-preserving guarantees. Many existing work of FL-based gradient boosting methods require different types of encryption-based protocols: homomorphic encryption \cite{homoxg1} \cite{Aono} \cite{homoxg3}, secret-sharing \cite{fang2020hybriddomain} and locality  sensitivity hashing \cite{li2019practical}, all of which result in significant communication/computation overhead. Applying \textit{Homomorphic Encryption (HE)} , \cite{Cheng} provides SecureBoost that offers high degree of privacy-preserving but requires high communication cost. Other approaches utilize \textit{Differential Privacy (DP)} and perform the analysis directly with the perturbed data as studied in \cite{li2021privacypreserving} , \cite{Shi2021}. Despite the reduction in training time, the model suffers accuracy loss by the injected noise. 
Our approach deviates from all the previous work. \textcolor{black}{We study a protocol that has lossless accuracy and achieves a compromise between model complexity and privacy-preserving}. We first formulate the evaluation of splitting score, the major step requiring privacy guarantee in XGBoost, as the multiplication of a categorical matrix and a vector, and then deploy a modified version of secure matrix multiplication (SMM) introduced in \cite{Karr}. \textcolor{black}{We show that if SMM is naively applied, due to its categorical entries the privacy guarantee of the matrix can be violated}.
We further point out the analogous scenario of privacy leakage that could be inherent in HE, yet neglected by the literature.  
To address the challenge, we enhance the SMM protocol to provide additional deniability and propose FedXGBoost-SMM, XGBoost for FL over vertically partitioned data with privacy-preserving guarantee.  In addition, utilizing noise perturbation with local differential privacy (LDP), we introduce FedXGBoost-LDP, which is a heuristic for practical perspective. Our contributions can be summarized as follows: 
\begin{itemize}
    \item \textbf{FedXGBoost-SMM:} a linear algebra based approach designed to achieve lossless model accuracy and more efficient in comparison to HE. \textcolor{black}{We vertically encode the categorical matrix and show that the extremely curious party can infer the true value of the matrix with low probability, while the overhead is negligible}. We provide modifications to enhance the privacy-preserving.
    \item \textbf{FedXGBoost-LDP}: a heuristically designed protocol with LDP that yields acceptable practical results. 
    \item \textbf{Practicality:} We experiment on real-world data and evaluate the utility of the heuristic FedXGBoost-LDP.
\end{itemize}

The remainder of the paper is organized as follows. 
Section \ref{sec:Pre} represents the preliminaries and the problem statement. Section \ref{sec:Main} introduces the applied privacy-preserving techniques and analyzes the potential information leakage of the proposed algorithms. Section \ref{sec:Prot} describes the procedures of FedXGBoost. The experiments and the evaluation of the protocols are provided in section \ref{sec:Exp}. Section \ref{sec:Con} concludes the study.

\section{Preliminaries} \label{sec:Pre}
\subsection{XGBoost - Gradient Tree Boosting}
\subsubsection{Regularized Learning Objective of Tree Boosting}
    Given a dataset $D$ = $\{(x_i, y_i), x_i \in \mathbb{X}, y_i \in \mathbb{R}\}$, $~x_i$ denotes feature vectors in feature space $\mathbb{X}$ and $y_i$ is the label of the $i^{th}$ instance. Let $n = |D|$ be the total amount of instances and $K$ be the amount of constructed regression trees. We have a regression model $ \phi(.)$ for an instance $x_j \in \mathbb{X}$ from multiple regression trees as follows
    \begin{equation}
        \hat{y}_j = \phi (x_j) = \sum_{k = 1}^K f_k(x_j), ~f_k \in \mathcal{F}, \\
    \end{equation}
    The set of regression trees $\mathcal{F}$ is defined as
    \[\mathcal{F} = \{f(x) = w_{q(x)}\}, ~ q: \mathbb{R}^m \xrightarrow{} T, w \in \mathbb{R}\]
    where $q$ denotes the tree structure that maps the instance to an unique leaf, $w$ is a weight of leaf, and $T$ is the amount of leaves of one tree. For any differentiable convex loss function $l: \mathbb{R}\times\mathbb{R} \rightarrow \mathbb{R}$, the objective function $\mathcal{L}(\phi)$ for the model training process is defined as
    \begin{equation} \label{eqn:costFuncL}
        \mathcal{L}(\phi) = \sum_{i = 1}^nl(\hat{y}_i, y_i) + \sum_k \Omega(f_k)\\
    \end{equation}
    in which the term $\Omega(f_k) = \gamma T + \frac{1}{2}\lambda\norm{w}^2$ is the regularization term to avoid over-fitting.
    
\subsubsection{XGBoost: Regression Tree Boosting}
    \cite{Chen} applied iterative optimization procedure to minimize the objective function (\ref{eqn:costFuncL}). At the $t^{th}$ iteration, new tree $f_t$ is constructed and contributes to the regression model. Therefore, the objective function at the $t^{th}$ iteration is formulated as
    \begin{equation} 
        \mathcal{L}^{(t)} = \sum_{i = 1}^n (l(y_i, \hat{y}^{(t-1)}_i + f_t(x_i))  + \Omega(f_t)\\
    \end{equation}
    The learning process aims to minimize the second order approximation of the objective function
    \begin{equation}  \label{eqn:optimizeFunc}
    \begin{split}
        &\min_{f_t} {\mathcal{L}}^{(t)} \\
         \approx &\min_{f_t} \sum_{i = 1}^n (l(y_i, \hat{y}^{(t-1)}) + g_i f_t(x_i) + \frac{h_i f^{2}_t(x_i)}{2}) + \Omega(f_t),\\
    \end{split}
    \end{equation}
    where $g_i = \partial_{\hat{y}_{(t-1)}}l$ and $~h_i = \partial^{2}_{\hat{y}_{(t-1)}}l$ are the first and  second derivative of the loss function at $\hat{y}_{(t-1)}$, respectively. Two factors that decide the efficiency of the training process are the structure of the new tree and the leaf weights. The tree is constructed by splitting the set of users into multiple nodes and analyzing their feature data. Many splitting options are proposed according to the feature distribution. Then the optimal candidate is selected to split the instances into left and right nodes. \cite{Chen} used the following equation to evaluate the splitting candidates
    \begin{equation} \label{eqn: LoptimalSplit}
         \begin{split}
             & \mathcal{L}_{split} = - \gamma + \\
             & \frac{1}{2}\left[ \frac{(\sum_{i \in I_L} g_i)^2}{\sum_{i \in I_L} h_i + \lambda} + \frac{(\sum_{i \in I_R} g_i)^2}{\sum_{i \in I_R} h_i + \lambda} - \frac{(\sum_{i \in I} g_i)^2}{\sum_{i \in I} h_i + \lambda} \right] 
         \end{split}
    \end{equation}
    where $I_L, I_R$ indicate the obtained left and right nodes by splitting the node $I$. Then, it continues to split from the new constructed nodes until reaching the maximum depth of the tree. The node at the last layer is the tree leaf that represents the weight for the common instance in this node. The leaf weight $w$ is equivalent to the prediction of a new tree and  contributes to the minimization of \eqref{eqn:optimizeFunc} through the term $f_t = w$. The optimal weight of each leaf is computed by
   \begin{equation} \label{eqn:optimalWeight}
        w^*_j = - \frac{\sum_{i \in I_j} g_i}{\sum_{i \in I_j} h_i + \lambda},
    \end{equation}
    where $I_j$ denotes the index set of instances that belong to the $j^{th}$ leaf, i.e., $I_j = \{i|q(x_i) = j\}$. As can be seen from (\ref{eqn: LoptimalSplit}) and (\ref{eqn:optimalWeight}), the tree construction requires the following knowledge:
\begin{itemize}
    \item The possible splitting options to construct new nodes, i.e.  $I_L, ~I_R$.
    \item The gradient and hessian values to evaluate the splitting options and compute the optimal weight. 
\end{itemize}

\subsection{Federated Learning over vertically partitioned data}
    The study by \cite{Qiang19} introduces various settings of FL. In this study, our proposed protocols focus on FL over vertically partitioned data, in which multiple databases own different features of the same sample instances. To clarify the settings, we introduce the concept of Active Party and Passive Party proposed by \cite{Cheng} through the following definition
    \begin{definition} \label{def:APPP}
    Active Party and Passive Party 
    \begin{itemize}
        \item \textbf{Active Party (AP)}: The party that holds both feature data and the class label. 
        \item \textbf{Passive Party (PP)}: The data provider party, which has only the feature data.
    \end{itemize}
    \end{definition}
    The protocol studied by \cite{Liang} is applied to determine the common database intersection between participants and align the features and class label with the corresponding sample ID securely. The main concern of XGBoost under this configuration is how to conduct the training process jointly between participants with the aligned database. 

\subsection{Secure Multi-party Computation (SMC)}
\cite{Cramer} defines SMC as techniques that allow multiple participants to compute accurately the final output without revealing private information. 
SMC protocols require the participants to follow a particular procedure that leads to the final result and guarantee under some assumptions that the private data can not be reconstructed. The SMC protocol applied in this paper is the Secure Matrix Multiplication (SMM) protocol motivated by \cite{Karr} as follows
{\protocol \textbf{Secure Matrix Multiplication by \cite{Karr}} \label{pro:Karr} \\
Let the parties A and B possess the private data matrix $D^A \in \mathbb{R}^{n \times m}$ and $D^B \in \mathbb{R}^{n \times l}$, respectively. They want to obtain the result $S = (D^A)^TD^B \in \mathbb{R}^{m \times l}$ without knowing the private information of the other participant. 
\begin{enumerate}
    \item Party A finds the set $\mathcal{U} = \{u_i \in \mathbb{R}^n| (D^A)^Tu_i = 0\}$, \textcolor{black}{which contains orthonormal null space vectors of $(D^A)^T$}. Then it selects $r$ kernel vectors to construct the matrix $Z = [u_1 \cdots u_r]\in \mathbb{R}^{n \times r}$. We have
    \[(D^A)^TZ = \textbf{0}^{m \times r}\]

    \item Party A sends the matrix $Z$ to party B, then party B computes the matrix $W \in \mathbb{R}^{n \times l}$ as 
        \[W = (I^{n \times n} - ZZ^T)D^B\]
      
    \item Party B sends $W$ back to party A to compute the true result of the multiplication by
        \[(D^A)^TW = (D^A)^T(I^{n \times n} - ZZ^T)D^B = (D^A)^TD^B\]
\qed
\end{enumerate}
}

 The designed protocol assumes that all participants are semi-honest, i.e., they honestly follow the protocol but they are curious about the private data of other participants. The private data can not be uniquely reconstructed from the data exchange of $Z, ~W$ in step 2 and 3 thanks to the rank deficiency of the linear equation systems.

We first formulate the challenge of XGBoost under FL settings as a SMC problem that can be solved by SMM in section \ref{smm_formulation}. Then we point out in section \ref{subsec:PPA1} and \ref{subsec:PPA2}  that the potential privacy threat if the introduced protocol is directly applied. Afterward, we propose our protocol, FedXGBoost-SMM, which enhances the degree of privacy-preserving.

\subsection{Differential Privacy (DP) \& Local Differential Privacy (LDP)}
    Different from SMC techniques, Differential Privacy by \cite{Dwork} guarantees privacy-preserving by introducing deniability in the private data. Informally, DP mechanisms inject calibrated noise into the query of the private dataset to make each individual's data indistinguishable. 
    {\definition{Differential Privacy, \cite{Dwork}} \\
    A randomized algorithm $\mathcal{M}$ satisfies $\epsilon$-differential privacy if for any data sets $x$ and $y$ differing on at most one element and all $\mathcal{S} \in \text{Range}(M)$:
    \[Pr(\mathcal{M}(x) \in \mathcal{S}) \leq exp(\epsilon)Pr(\mathcal{M}(y) \in \mathcal{S})\]
    }
    A stronger DP approach is Local Differential Privacy, which perturbs the individual's data locally to protect the private information.
    {\definition{Local Differential Privacy, \cite{Erlingsson}} \\
    The perturbation mechanism $\pi(.)$ satisfies $\epsilon$- local differential privacy if for any two input $t, ~t'$ in the domain of $\pi(.)$ and any output $t^*$ in the range of $\pi(.)$, there is an $\epsilon > 0$ that
    \[Pr[\pi(t) = t^*] \leq exp(\epsilon)Pr[\pi(t') = t^*]\]
    }
    Many efficient DP and LDP perturbation mechanisms that enable statistical analysis of perturbed data are studied in \cite{Duchi}, \cite{Wang}. In comparison to SMC techniques, LDP methods offer efficient computational costs because the perturbed data can be used directly in the analysis. 
    However, due to the injected noise, the model accuracy is negatively impacted. This motivates us to design FedXGBoost-LDP, which is a heuristic approach that finds the compromise between model complexity and accuracy. 
    
    {\remark{The two protocols, FedXGBoost-SMM and FedXGBoost-LDP, are independent of each other. FedXGBoost-SMM is a theoretical approach applied linear algebra techniques to achieve the lossless property. FedXGBoost-LDP is a heuristic approach that conducts regression training with noise perturbed data. In principle, FedXGBoost-LDP is faster, and with small theoretical modification to increase the accuracy.}}
    
\section{Main Results} \label{sec:Main}
    To construct the new tree nodes, passive parties first (PP) analyze their user's distribution according to their feature data. They they propose splitting candidates that separate the current node (the set of users being analyzed) into left and right nodes. The distribution analysis over a large data set is executed by the efficient Approximate Quantile algorithms as studied in \cite{Karnin}, \cite{Ping08}, \cite{Stephen11}. Afterward, the optimal splitting candidate is determined by comparing the loss reduction between splitting candidates as shown in (\ref{eqn: LoptimalSplit}). For brevity, in XGBoost under FL settings, PP plays a role as a splitting candidate owner. On the other side, the active party with the private class label owns the confidential gradient and hessian values of users.
    During the regression learning, AP and PP desire to compute the optimal splitting candidate securely.
    \textcolor{black}{\remark \label{rem:AQ} The description of the Approximate Quantile algorithm to analyze the users' distribution and propose splitting candidates are introduced in \cite{Chen}. Note that we apply the trivial Quantile algorithm instead of the Weighted Quantile Sketch.}
    
    We introduce the concept of the splitting operator and splitting matrix to formulate the private information of the passive parties as follows
    \subsection{Splitting Matrix - Private Data of Passive Parties}
    {We simplify the notations during the formulation by considering only one AP that has the true label of users, and one PP that constructs the splitting matrix based from one feature. The complete procedure for multiple PP with many features are described in section \ref{sec:Prot}.}
    
    {\definition{Splitting Operator and Splitting Matrix \\} 
    Let $\mathbb{X}$ be the feature space, $f^k = [x_1 ~ x_2 \cdots x_n]^T \in \mathbb{X}^n$ be the values of the $k^{th}$ feature of $n$ users, and $\mathcal{S} = \{s_1,s_2,\cdots,s_l\} \subseteq \mathbb{X}^l$ be the set of $l$ splitting candidates. The splitting operator $\text{Split}(f^k,\mathcal{S}): \mathbb{X}^{n} \times \mathbb{X}^l \longrightarrow \{0,1\}^{n \times l}$ performs the splitting operation by comparing all feature data with all splitting candidates and outputs the splitting matrix $M \in \{0,1\}^{n \times l}$ as \\
    \begin{center}
    $M = \text{Split}(f^k,\mathcal{S})$ =  $\begin{pmatrix}
                    u_{1s_1}  & u_{1s_2} & \cdots & u_{1s_l} \\
                    \vdots  & \vdots & \vdots & \vdots \\
                    u_{ns_1}  & u_{ns_2} & \cdots & u_{ns_l} \\
                    \end{pmatrix}$,
    where $u_{is_j} = \twopartdef{1}{, ~x_i \leq s_{j}}{0}{, ~x_{i} > s{j}}$
    \end{center}
    }
    
    \begin{figure} [tb]
        \centering  
        \includegraphics[width=90mm]{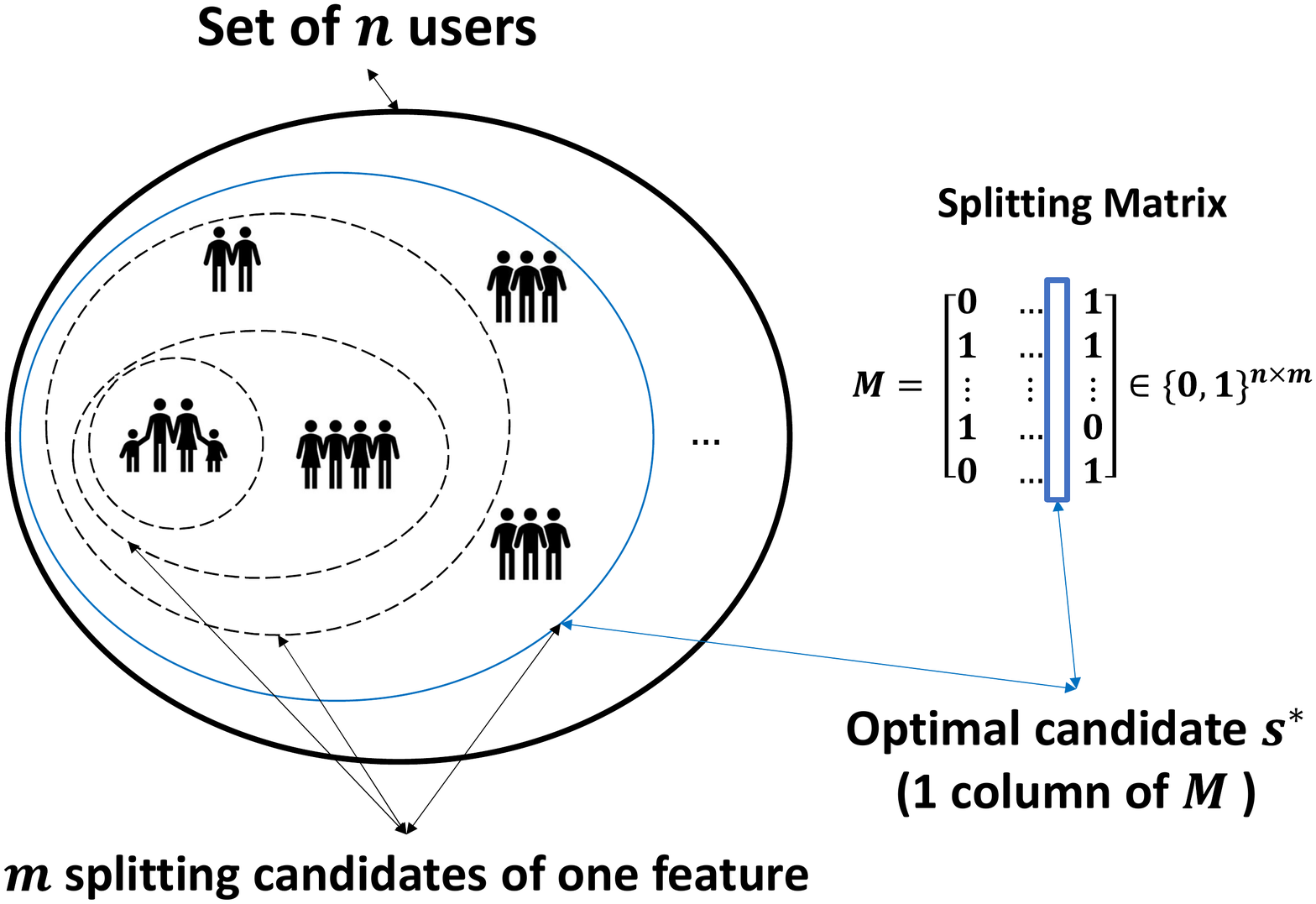}
        \caption{Construction of a splitting matrix}
        \caption*{\textcolor{black}{The passive party analyzes the feature distribution and split the set of users into two subsets. Each circle (dash-dotted line) represents one splitting candidates and corresponds to one column of the splitting matrix $M$.}} 
        \label{fig:SMProp}
    \end{figure}
        
    Each column of the splitting matrix represents one splitting candidate, which labels any user as "1" if they belong to the left node and "0" for the right node. Figure \ref{fig:SMProp} and the following example depicts the functionality of the splitting operator and the splitting matrix.
    {\example{} 
    Let the feature vector be $f^k = [1,~11,~9,~4,~2,~12,~17,~13,~5]^T$. The PP proposes a set of three splitting candidates $\mathcal{S} = \{s_1 = 11, ~s_2 = 6, ~s_3 = 12\}$. The result of the splitting operator applied to the feature vector $f^k$ and the candidate set $\mathcal{S}$ is the following splitting matrix $M$
    \begin{center}
        $M$ = \text{Split}($f^k$,$\mathcal{S}$) = $\begin{pmatrix}
                            1 ~1 ~1 ~1 ~1 ~0 ~0 ~0 ~1 \\
                            1 ~0 ~0 ~1 ~1 ~0 ~0 ~0 ~1 \\
                            1 ~1 ~1 ~1 ~1 ~1 ~0 ~0 ~1 \\
                                        \end{pmatrix}^T$   
    \end{center}
    }
    According to (\ref{eqn: LoptimalSplit}), the evaluation of each splitting candidates require the aggregated gradients and hessians of instances in the left and right nodes. These are computed through the multiplication of the splitting matrix and the data vector. Particularly, let $i \in \{1 \cdots n\}$ denote the user index, $\mathcal{N}$ indicate the set of all users being analyzed and $n = |\mathcal{N}|$. The two subsets $\mathcal{N}_L, \mathcal{N}_R$ are obtained by the splitting operator, such that $\mathcal{N}_L \cap \mathcal{N}_R = \varnothing, ~\mathcal{N}_L \cup \mathcal{N}_R = \mathcal{N}$. Let the gradient and hessian vector of $n$ users with numerical values be
    \[g = (g_{1} \cdots g_{n})^T, h = (h_{1} \cdots h_{n})^T \in \mathbb{R}^n\]
    From the set $\mathcal{S} =  \{s_1, \cdots, s_l\}$ that represents $l$ splitting candidates, we use the index $s_i$ to indicate one particular candidate. The aggregated gradients and hessians of users in the left and right nodes, which are $G^L$, $H^L$ and $G^R$, $H^R$ respectively, are computed from $M, g, h$ as follows
    \begin{equation} \label{eqn:computeSumGH}
        \begin{split}
            G & = \sum_{i \in \mathcal{N}}g_i \in \mathbb{R}, ~H = \sum_{i \in \mathcal{N}}h_i \in \mathbb{R} \\
            G^L & = (G^L_{s_1} ~\cdots ~G^L_{s_l})^T = M^Tg \in \mathbb{R}^l \\
            H^L & = (H^L_{s_1} ~\cdots ~H^L_{s_l})^T = M^Th \in \mathbb{R}^l \\
            G^R & = (G^R_{s_1} ~\cdots ~G^R_{s_l})^T = G.\textbf{1}^l - G^L \in \mathbb{R}^l\\
            H^R & = (H^R_{s_1} ~\cdots ~H^R_{s_l})^T = H.\textbf{1}^l - H^L \in \mathbb{R}^l \\
        \end{split}
    \end{equation}
    where $G^L_{s_i}, G^R_{s_i}, H^L_{s_i}, H^R_{s_i}$ represent the aggregated gradients and hessians of the left and right node split by \textbf{one splitting candidate $s_i$}. After obtaining the result of the matrix multiplication, each candidate $s_i$ is then evaluated by comparing the splitting score $\mathcal{L}^{s_i}_{split}$ from (\ref{eqn: LoptimalSplit}). The best candidate $s^*$ has the highest splitting score. \\
    \begin{equation} \label{eqn:Lsplit_short}
    \begin{split}
        s^* &= \arg \max_{s_i \in \mathcal{S}} \mathcal{L}^{s_i}_{split} \\
         & = \arg \max_{s_i \in \mathcal{S}} \frac{1}{2}\left[ \frac{(G^L_{s_i})^2}{(H^L_{s_i}) + \lambda} + \frac{(G^R_{s_i})^2}{(H^R_{s_i}) + \lambda} - \frac{G^2}{H + \lambda} \right] - \gamma
    \end{split}
    \end{equation}
    If a splitting operator constructs the last layer of the tree, i.e., it constructs the tree leaves, the optimal weight for each leaf is computed by (\ref{eqn:optimalWeight}) as
    \begin{equation} \notag 
    \begin{split}
        w^*_L & = - \frac{G^L_{s^*}}{H^L_{s^*} + \lambda}, \quad w^*_R = -\frac{G^R_{s^*}}{H^R_{s^*} + \lambda}
    \end{split}
    \end{equation}
    
    {\remark \textcolor{black}{The previous studies use the addition operation to compute the aggregated gradients and hessians between multiple parties. We instead introduce the splitting matrix to mathematically formulate the private data and the functionality of the PP.}}  \\
  
    
    \subsection{Problem formulation - The relation between \\FedXGBoost and Secure-Matrix Multiplication}
    \label{smm_formulation}
    \begin{figure} [htbp]
    \centering  
    \includegraphics[width=90mm]{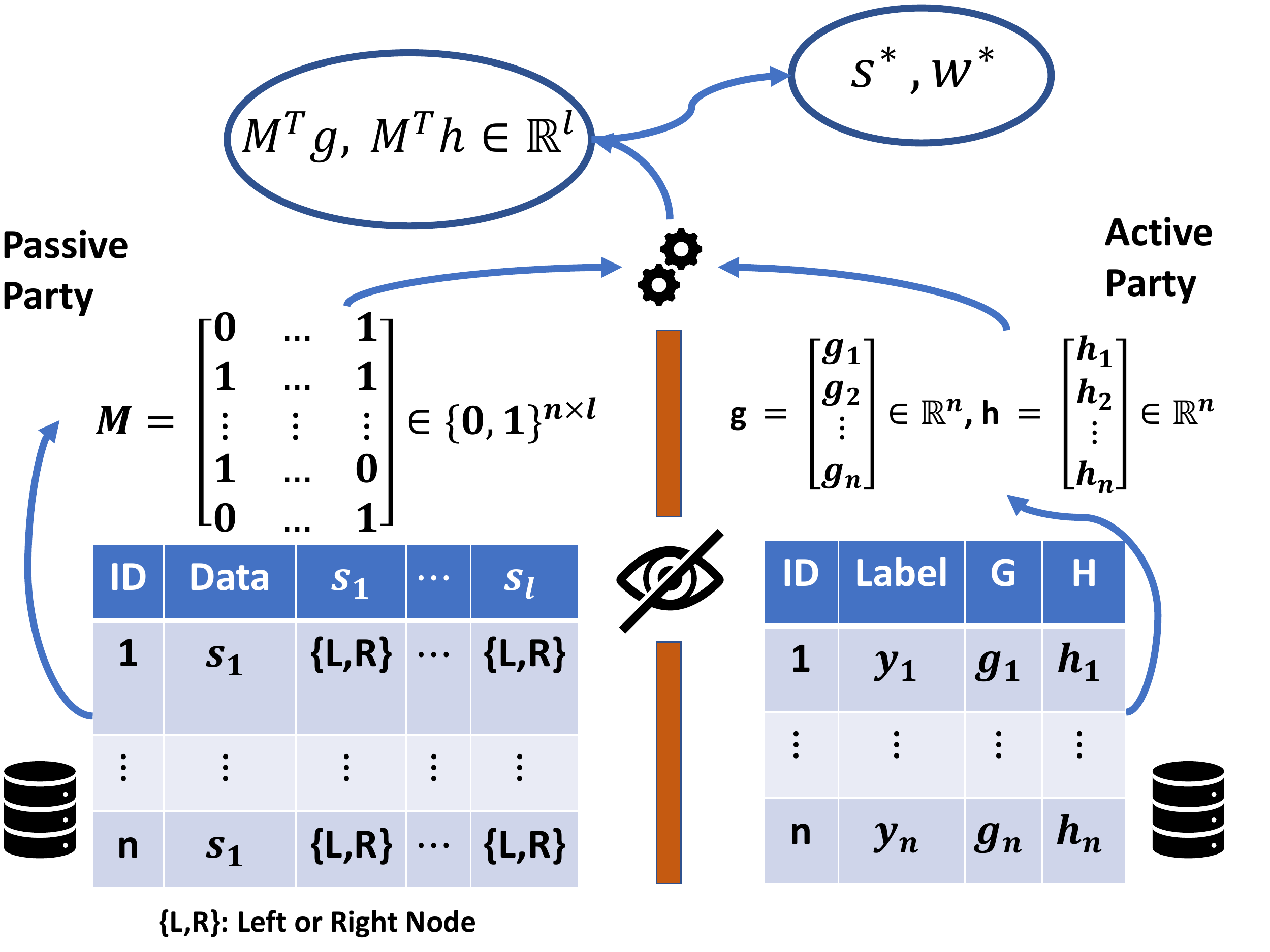}
    \caption{From Federated Learning XGBoost to SMM}
    \label{fig:SMM}
    \end{figure}
    
    From the concept of the splitting matrix, we reformulate the challenges of FedXGBoost into an SMC problem that can be solved by SMM protocols.
    Particularly, according to (\ref{eqn:computeSumGH}), the AP has $g,~h$, while the PPs own $M$ and they desire to compute $M^Tg, ~M^Th$ securely. Then the results are applied to  (\ref{eqn:computeSumGH}) and (\ref{eqn:Lsplit_short}) to find the best splitting candidate and the optimal leaf weights. Figure \ref{fig:SMM} illustrates the concept. The next subsections use the introduced notations of $M \in \{0,1\}^{n \times l}, g,h \in \mathbb{R}^n$ to design two secure protocols, which are FedXGBoost-SMM and FedXGBoost-LDP, and analyze the information leakage of these two protocols. 
    {\remark The two protocols FedXGBoost-SMM \& FedXGBoost-LDP are independent and only different in their applied privacy-preserving techniques that allow the splitting scores to be evaluated securely. Therefore, we focus on analyzing this most crucial step, which determines the best splitting candidate. Additionally, the protocols are designed for semi-honest participants. 
    }

    \subsection{FedXGBoost-SMM}
    FedXGBoost-SMM is motivated by the SMM protocols that allow the participants to determine the optimal splitting candidate securely. We design two versions of FedXGBoost-SMM procedures that consider the different roles of AP and PP according to Protocol \ref{pro:Karr}. The first version (FedXGBoost-SMM-1) considers $D^A$ as the gradient and the hessian $g, ~h$ vectors of the AP or combined as a matrix $[g ~h] \in \mathbb{R}^{n \times 2}$, while $D^B$ is the splitting matrix of PP. Conversely, in the second version (FedXGBoost-SMM-2), $D^A$ represents the splitting matrix of PP and $D^B$ indicates $[g ~h]$ of AP. 
    {\remark \textcolor{black}{Even though the roles of parties are not important in the original SMM protocol, we point out in the privacy-leakage analysis that the potential threat of FedXGBoost-SMM-1 and FedXGBoost-SMM-2 are different. In general, FedXGBoost-SMM-1 achieves equivalent privacy-preserving as SecureBoost in \cite{Cheng} using homomorphic encryption technique. Nevertheless, under the assumption that any extreme curious party has  unbounded computational capability, the private data of PPs in both SecureBoost and FedXGBoost-SMM-1 can be reconstructed. Motivated by this, the designed FedXGBoost-SMM-2 prevents the private data from being inferred with high confidence. This explains why FedXGBoost-SMM-2 requires sophisticated privacy-preserving but results in higher complexity. \\
    }}

    \subsubsection{FedXGBoost-SMM-1 }
    \paragraph{Protocol description}
    Algorithm \ref{alg:SMFedXGBoost-1} describes the procedure of FedXGBoost-SMM-1. AP first determines the set of users being analyzed. Then it announces this set to all PP and  transmits the generated orthonormal null-space vectors of $[g ~h]^T \in \mathbb{R}^{n \times 2}$. Each PP analyzes the feature of the announced user set and constructs its private splitting matrix. Then it applies Algorithm \ref{alg:Secure-Response} to generate a secure response $W \in \mathbb{R}^{n \times l}$. 
    The AP requests $W$ from all PP to compute the aggregated gradients and hessians of each splitting candidate, which are the elements of $W^Tg, ~W^Th \in \mathbb{R}^{l}$. Afterward, it computes the splitting score and finds the optimal score between all PP. The PP with optimal score is requested to reveal the corresponding splitting operation. The AP then constructs new nodes and repeats the process with the new set of users.
      
    \begin{algorithm} [tb]
    \caption{FedXGBoost-SMM-1}
    \label{alg:SMFedXGBoost-1}
    \textbf{Input:}
    \begin{itemize}
        \item Active party (\textbf{AP}) has $g, ~h \in \mathbb{R}^n$
        \item Total $p$ passive parties ($\textbf{PP}$), $k^{th}$ PP has splitting matrix $M^k \in \{0,1\}^{n \times l}$
    \end{itemize}
    \textbf{Output of protocols:} 
    The optimal splitting operation \\
    \textbf{Procedures:}
    \begin{algorithmic}[1] 
    \STATE $\textbf{AP}:$ Find the set orthonormal null-space vectors of $[g ~h]^T$: $~\mathcal{U} \leftarrow \{u_i \in \mathbb{R}^n| [g ~h]^T(u_i) = 0\}, ~|\mathcal{U}| = r$
    \STATE $\textbf{AP}:$ Transmit $~\mathcal{U}$  to all PP
    \STATE $\textbf{PP}_k$:  $W^k \in \mathbb{R}^{n \times l} \xleftarrow{Algo. \ref{alg:Secure-Response}}$ Secure-Response($M^k$, $~\mathcal{U}$)  
    \STATE $\textbf{PP}_k:$ Transmit $W^k$ to AP
    \STATE $\textbf{AP}: ~G \leftarrow \sum_{i = 1}^n g_i, ~H \leftarrow \sum_{i = 1}^n h_i$
    \STATEx /* AP finds optimal score from all PP */
    \STATE $(L^k)^* \leftarrow - \infty, ~(s^k)^* \leftarrow 0$
    \FOR{k = 1 \textbf{to} p}
    \FOR{i = 1 \textbf{to} l}
    \STATE $\{~G^L_{s_i}, H^L_{s_i}\} \leftarrow \{(W^k)^Tg\}_i, \{(W^k)^Th\}_i$ 
    \STATE $G^R_{s_i} \leftarrow G - G^L_{s_i}, ~H^L_{s_i} \leftarrow H - H^L_{s_i}$ \
    \STATE $L \leftarrow \frac{1}{2} \left[ \frac{(G^L_{s_i})^2}{H^L_{s_i} + \lambda} + \frac{(G^R_{s_i})^2}{H^R_{s_i} + \lambda} - \frac{G^2}{H + \lambda}\right]$
    \IF{$(L^k)^* < L$}
    \STATE $(L^k)^* \leftarrow L, ~(s^k)^* \leftarrow i$
    \ENDIF
    \ENDFOR
    \ENDFOR
    \STATE \textbf{AP:} $k^* = \operatorname*{argmax}_{k \in \{1 \cdots p\}} L^k, ~(s^k)^*$
    \end{algorithmic}
    \textbf{Return:} $\text{PP}_{k^*}$ and its optimal splitting operation
    \end{algorithm}
     
    \begin{algorithm} [tb]
    \floatname{algorithm}{Algorithm} 
    \caption{Secure-Response}
    \label{alg:Secure-Response}
    \textbf{Input:} 
    \begin{itemize}
        \item Private data $X \in \mathbb{R}^{n \times p}$
        \item Received $\mathcal{U} = \{u_i \in \mathbb{R}^n, i \in \{1, \cdots, r\}\}$
    \end{itemize}
    \textbf{Output:} $W \in \mathbb{R}^{l \times n}$ \\
    \textbf{Procedures:} 
    \begin{algorithmic}[1]
    \STATE Select random $r'$ vector in $\mathcal{U}$, with $r' \leq r$ 
    \STATE $Z \leftarrow [u_1 \cdots u_{r'}] \in \mathbb{R}^{n \times r'}$
    \STATE $W \leftarrow (I^{n \times n} - ZZ^T)X \in \mathbb{R}^{n \times p}$
    \end{algorithmic}
    \textbf{Return:} $W$
    \end{algorithm}
    
    \paragraph{Analysis of privacy-preserving} \label{subsec:PPA1}
    The study of SecureBoost provides a thorough analysis of potential privacy leakage for the AP. In our study, we focus on the private splitting matrix of the PPs. Despite these matrices do not contain the real feature data, they reveal the user's distribution. Such  information allows the curious party to infer the range of feature values. The potential privacy leakage of FedXGBoost-SMM-1 are caused by 1) The revealed null-space vectors set $\mathcal{U}$ of AP; 2) The response $W$ of PP; 3) AP knows the aggregated gradients and hessians of all splitting candidates; 
    \begin{enumerate}
        \item \emph{PP knows the null-space vectors of AP} \\ 
        Let the set of $r$ null-space vectors construct a matrix $U \in \mathbb{R}^{n \times r}$, with $rank(U) = r, ~ r \leq n-2$, reconstructing $[g ~h]$ is equivalent to the following problem
        \begin{problem}
        Find $x \in \mathbb{R}^{n \times 2}$ with a given $U \in \mathbb{R}^{n \times r}, ~\text{rank}(U) = r$ that satisfies
        \begin{equation} \label{eqn:nsgh}
        U^Tx = 0^r \in \mathbb{R}^{r}
        \end{equation}
        \end{problem}
        There exist infinite solutions for (\ref{eqn:nsgh}) due to the rank deficiency of the linear equation system. The span of $x$ can be inferred if $r \approx n-2$. However, it is sufficient to choose $1 \ll r \ll n-2$  
        \item \emph{AP knows the response $W$ from PP} \\
        As described in the Algorithm \ref{alg:SMFedXGBoost-1}, PP randomly selects $r'$ vectors in $\mathcal{U}$ to construct $Z$, with $r' < r$. Then it computes
        \begin{equation}
        W = (I^{n \times n} - ZZ^T)M \in \mathbb{R}^{n \times l}
        \end{equation}
        {\property{\cite{Karr} \label{prop:noninv} The constructed Z from the received null-space vectors satisfies $rank(I^{n \times n} - ZZ^T) < n$ so it is not invertible.}} \\
        In the original protocol, \cite{Karr} stated that the information of $rank(W)$ contributes to the privacy leakage. For this reason, the randomness introduced in Algorithm \ref{alg:Secure-Response} conceals the information of $ZZ^T$. 
        
        \item \emph{AP knows the aggregated gradients and hessians of all splitting candidates} \\
        Consider the AP is curious about $M$. The effort to reconstruct possible splitting candidates is equivalent to the following integer programming problem
        \begin{problem}\label{prob:secureZ1}
        \textit{Find $\{x_1, ~\cdots, x_n\} \in \{0,1\}$ such that for a given $A = (a_1 ~\cdots ~a_n)^T \in \mathbb{R}^{n}$ and $b \in \mathbb{R}$ it satisfies
        \begin{equation} \notag
        \sum_{i = 1}^n x_ia_i - b = 0 
        \end{equation}
        }
        \end{problem}
       From our understanding, Problem \ref{prob:secureZ1} belongs to the set of NP-complete problems. This guarantees the privacy-preserving under the assumption of bounded computational capability. \emph{This challenge also occurs in methods applying homomorphic encryption techniques, yet was not mentioned in the previous literature}. In SecureBoost, AP encrypts the gradients and hessians as $\left<g\right>\footnote{$\left <.\right>$ operator indicates encrypted data} , ~\left<h\right>$ before transmitting these to PP. \emph{Each PP aggregates the encrypted gradients and hessians and sends back to the AP, i.e., PP computes the multiplication of the splitting matrix and the encrypted vectors to obtain the encrypted $\left<M^Tg\right>$ and $\left<M^Th\right>$, respectively.} 
        {\observation \label{obs:NP1} If there exists an efficient algorithm that solves Problem \ref{prob:secureZ1} in polynomial time, the splitting matrix can be reconstructed from the known aggregated gradients $\left<M^Tg\right>$ and hessians $\left<M^Th\right>$.}\\
    \end{enumerate}
        To this end, we conclude that FedXGBoost-SMM-1 achieves equivalent privacy-preserving as SecureBoost that applies homomorphic encryption techniques. The open question from Observation \ref{obs:NP1} motivates us to design FedXGBoost-SMM-2.

    \subsubsection{FedXGBoost-SMM-2}
    \paragraph{FedXGBoost-SMM-2 - Protocol description}
    Algorithm \ref{alg:SMFedXGBoost-2} describes the procedure of FedXGBoost-SMM-2.
    From a given set of users, the PP analyzes the feature distribution and construct the splitting matrix $M \in \{0,1\}^{n \times l}$. Then the PP applies Algorithm \ref{alg:Secure-Kernel} to generate 
    a quasi-secure splitting matrix $M^* \in {\mathbb{R}^{n \times l'}}$ and a set $\mathcal{U}$ that contains $r$ null-space vectors of $(M^*)^T$, i.e., $~\mathcal{U} = \{u_i \in \mathbb{R}^n| (M^*)^Tu_i = 0\}, ~|\mathcal{U}| = r$. \emph{Algorithm  \ref{alg:Secure-Kernel} tackles the potential privacy leakage against curious party that has unbounded computational capability}. The set $\mathcal{U}$ is then transmitted to the AP. The AP applies Algorithm \ref{alg:Secure-Response} to construct a secure response $W \in \mathbb{R}^{n \times 2}$. 
    The PP requests $W$ from AP and computes $(M^*)^TW \in \mathbb{R}^{l' \times 2}$. Then it applies a post-processing step to obtain $(M^T)W \in \mathbb{R}^{l \times 2}$ that contains the aggregated gradients and hessians of each splitting candidate in each column respectively. To this end, the PP computes the splitting score of $l$ candidates according to (\ref{eqn: LoptimalSplit}) and determines the optimal one. Afterward, the PP sends the optimal score to the AP, which will be evaluated between multiple PPs. If the splitting option of each PP is selected, AP requests the splitting operation of that candidate (the specific column of the private splitting matrix), constructs new nodes, and repeats the process.


    \begin{algorithm} [H]
    \caption{FedXGBoost-SMM-2}
    \label{alg:SMFedXGBoost-2}
    \textbf{Input:}
    \begin{itemize}
        \item Users set $\mathcal{N}_I$, with $|\mathcal{N}_I| = n$ 
        \item Active party (\textbf{AP}) has $g, ~h \in \mathbb{R}^n$
        \item Total $p$ passive parties ($\textbf{PP}$), $k^{th}$ PP has data base $D^k$
    \end{itemize}
    \textbf{Output of protocols:} 
    The optimal splitting operation \\
    \textbf{Procedures:}
    \begin{algorithmic}[1] 
    \STATEx \text{/* Each passive party finds its best splitting candidate */}
    \FOR{k = 1 \textbf{to} p}
    \STATE $\textbf{PP}_k:$
    \STATEx$~~~~~~~\{\mathcal{U}^k, M^*, \mathcal{S}, \mathcal{V}\} \xleftarrow{Algo. \ref{alg:Secure-Kernel}}$ Secure-Kernel($D^k$,$~\mathcal{N}_I$)
    \STATE $\textbf{PP}_k:$ Transfer $\mathcal{U}^k$ to AP.
    \STATE $\textbf{AP}:$
    \STATEx$~~~~~~~\{W, G, H\}\xleftarrow{Algo. \ref{alg:Secure-Response}}$ Secure-Response($[g ~h]$, $~\mathcal{U}^k$)
    \STATE \textbf{AP:} Transfer \{$W, G, H$\} to PP$_k$.
    \STATE $\textbf{PP}_k:$ 
    \STATEx /* Compute the aggregated gradients and hessians */ 
    \STATE $W_g, ~W_h \leftarrow W$ $~~~~~~~~~~~ \text{/*} W = [W_g ~ W_h] \in \mathbb{R}^{n \times 2}$ */
    \STATE $\{s_i, ~G^L_{s_i}, H^L_{s_i}\} \leftarrow \{i, ~\{(M^*)^T W_g\}_i, ~\{(M^*)^T W_h\}_i\}$ 
    \STATE $(L^k)^* \leftarrow - \infty, ~(s^k)^* \leftarrow 0$
    \STATEx \text{/* Post-processing: Consider only results with index in $\mathcal{V}$ */}
    \FOR{i $\in \mathcal{V}$} 
    \STATE $G^R_{s_i} \leftarrow G - G^L_{s_i}, ~H^L_{s_i} \leftarrow H - H^L_{s_i}$ \
    \STATE $L \leftarrow \frac{1}{2} \left[ \frac{(G^L_{s_i})^2}{H^L_{s_i} + \lambda} + \frac{(G^R_{s_i})^2}{H^R_{s_i} + \lambda} - \frac{G^2}{H + \lambda}\right]$
    \IF{$(L^k)^* < L$}
    \STATE $(L^k)^* \leftarrow L, ~(s^k)^* \leftarrow s_i \in \mathcal{S}$
    \ENDIF
    \ENDFOR
        \STATE Transfer $(L^k)^*$ to AP, save $(s^k)^*$
    \ENDFOR 
    \STATEx \text{/*AP finds the best splitting candidate over $p$ parties*/}
    \STATE \textbf{AP:} 
    \STATEx Find $k^* = \operatorname*{argmax}_{k \in \{1 \cdots p\}} L^k$, $~(s^k)^*$
    \end{algorithmic}
    \textbf{Return:} $\text{PP}_{k^*}$ and its optimal splitting operation
    \end{algorithm}

    \begin{algorithm} [tb]
    \caption{Secure-Kernel}
    \label{alg:Secure-Kernel}
    \textbf{Input:} 
    \begin{itemize}
        \item Users set $\mathcal{N}_I$, with $|\mathcal{N}_I| = n$
        \item $D^k = (f_1 ~f_2 \cdots f_d)^T \in \mathbb{R}^{n \times d}$: Database contains $d$ feature data of $n$ users.
    \end{itemize}
    \textbf{Output:}
    \begin{itemize}
        \item $\mathcal{U}$: Set $r$ of null-space vectors of $M^*$
        \item $M^* \in \mathbb{R}^{n \times l'}$: Quasi-secure splitting matrix
        \item $\mathcal{S}$: Set of the proposed splitting candidates
        \item $\mathcal{V}$: Index list of the true splitting candidates $M$ in $M^*$ 
        \end{itemize}
    \textbf{Brief Explanation:}
    \begin{itemize}
        \item Random splitting matrix $M'$ cover $M$ in $M^*$
        \item Matrix $Y$ with random numerical values handles the sparsity of $M^*$ to construct the usable null-space vectors
    \end{itemize}
    \textbf{Procedures:} 
    \begin{algorithmic}[1]
    \FOR{j = 1 \textbf{to} d} 
    \STATEx /* Propose splitting candidates for each feature */
    \STATE $S_j \leftarrow$ Approximate Quantile($f_j$) /* Remark \ref{rem:AQ} */
    \STATEx /* Construct the splitting matrix */
    \STATE $M_j \leftarrow$ \text{Split}$(f_j, S_j)$ 
    \ENDFOR
    \STATE $M  \leftarrow [M_1 ~M_2 \cdots ~M_d] \in \{0,1\}^{j \times l}$ /*  Merge feature */
    \STATE $M'\leftarrow$ random $\{0,1\}^{n \times l_1} $ 
    \STATE $Y \leftarrow \mathcal{N}(\mu, \sigma) \in \mathbb{R}^{n \times l_2}$ /* Remark \ref{rem:explainY} */
    \STATEx /* Construct the quasi-secure splitting matrix */
    \STATE $M^* \leftarrow [M ~M'  ~Y] \in \mathbb{R}^{n \times l'}, ~l' = l + l_1 + l_2$
    \STATE $\mathcal{V} \leftarrow$  Index of $M$ in $M^*$  
    \STATEx /* Construct set of $r$ null-space vectors of $(M^*)^T$ */
    \STATE $~\mathcal{U} \leftarrow \{u_i \in \mathbb{R}^n| (M^*)^Tu_i = 0\}, ~|\mathcal{U}| = r$
    \end{algorithmic}
    \textbf{Return:} $\mathcal{U}, ~M^*, ~\mathcal{S}, ~\mathcal{V}$
    \end{algorithm}

    \paragraph{Analysis of privacy-preserving} \label{subsec:PPA2}

\begin{figure} [tbp]
        \centering  
        \includegraphics[width=90mm]{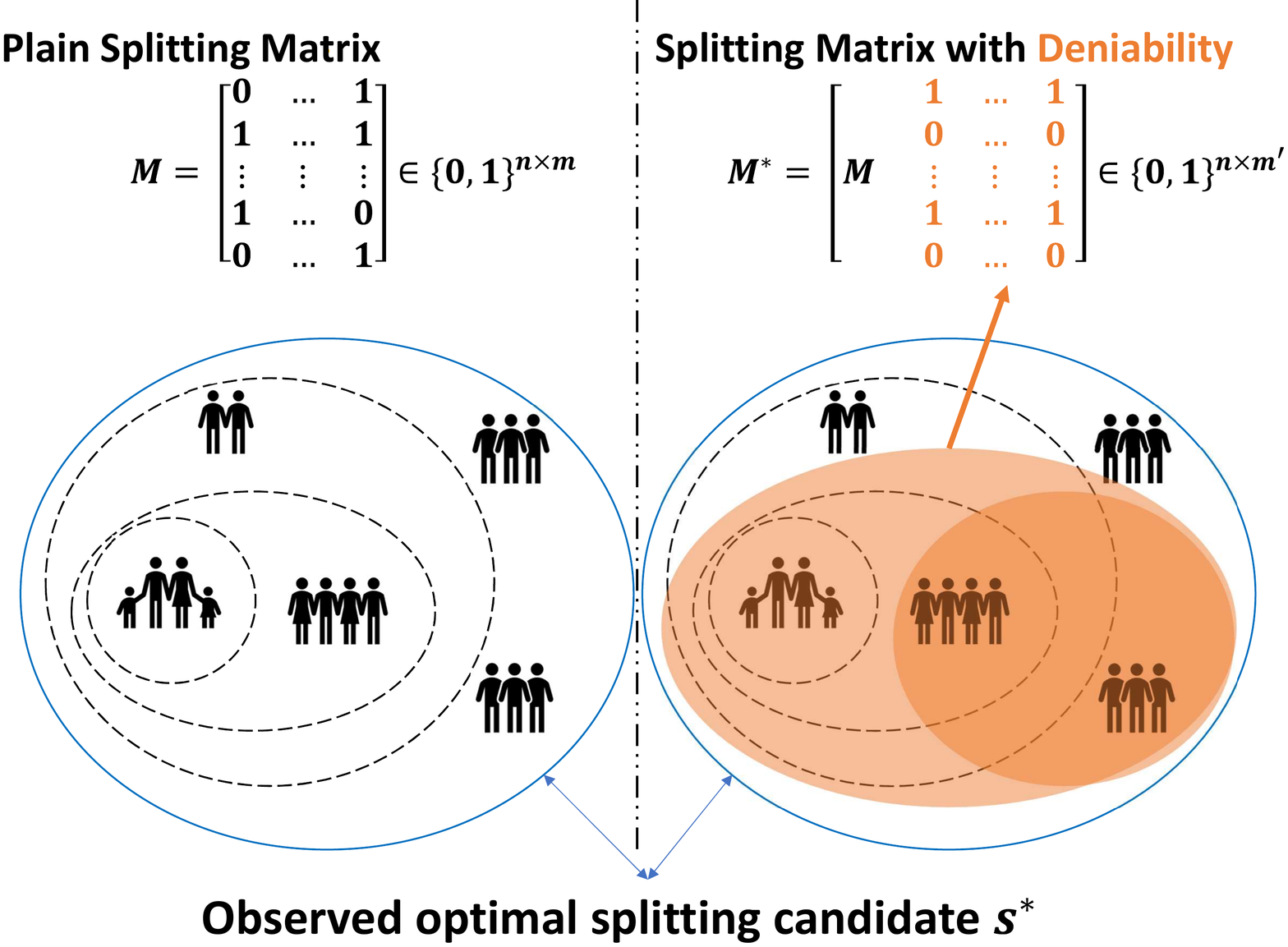}
        \caption{Splitting matrix with injected random columns}
        \caption*{\textcolor{black}{Adding random splitting candidates, i.e, adding random $\{0,1\}$ columns into the real splitting matrix prevents users' labels from being inferred with high probability.}}
        \label{fig:SMPerturb}
        \end{figure}

    The potential leakages are caused by ~1) Algorithm \ref{alg:Secure-Kernel}: The null-space of the splitting matrix; ~2) The optimal splitting candidate (one private column of the splitting matrix); ~3) Algorithm \ref{alg:Secure-Response}: The result $W$ that PP requests from AP; 
    We analyze the potential privacy leakage of the proposed FedXGBoost-SMM through the following mathematical arguments
    \begin{enumerate}
        \item \emph{AP knows the set of $r$ null-space vectors of matrix $M^*$ generated by Algorithm \ref{alg:Secure-Kernel}}. \\ 
        This potential leakage is similar to Problem \ref{prob:secureZ1} and it is the main motivation for FedXGBoost-SMM-2. We enhance the privacy of the splitting matrix by Algorithm \ref{alg:Secure-Kernel}. This algorithm adds calibrated random $\{0,1\}$ columns before constructing the null-space, i.e.,
        \[M^* = [M ~M'] \in \{0,1\}^{n \times (l + l_1)}\],
        where $M' \in \{0,1\}^{n \times l_1}$ is properly generated.
        If there exists an efficient algorithm as declared in the Observation \ref{obs:NP1}, the private splitting label can not be inferred with high confidence. Figure \ref{fig:SMPerturb} illustrates the concept of Algorithm \ref{alg:Secure-Kernel}. 
        Note that these randomness do not affect the final result because the columns are independent from each other. The result from generated columns are omitted in the evaluation of splitting candidates. We conclude that the private splitting matrix is secured through the generated enhanced null-space.
        
        \item \emph{AP knows the optimal splitting option.} \\
        After the optimal splitting score is determined, AP requests the corresponding optimal splitting operation, which is one column of the private splitting matrix $M$ from the selected PP. Despite this column does not reveal any additional private information, it is the main reason that requires the random $\{0,1\}$ columns in Algorithm \ref{alg:Secure-Kernel} to be generated properly. 
        
       \item \emph{PP knows the $W$ generated by Algorithm \ref{alg:Secure-Response} from the AP.} \\
       In step 2 of the original SMM Protocol \ref{pro:Karr}, the AP computes $W$ from the obtained $Z$ from the PP as
       \begin{equation} \label{secureW}
       W =(I^{n \times n} - ZZ^T) V \in \mathbb{R}^{n \times 2},
       \end{equation}
       where $V = [g ~h] \in \mathbb{R}^{n \times 2}$ in our scenario. 
       As guaranteed by Property \ref{prop:noninv}, there exist infinite solutions for (\ref{secureW}). However, if $ZZ^T$ is sparse that has rows with one element different from zero, the true values of $V$ for some users are observable. Indeed, this is a critical problem due to the property of the null-space of a categorical matrix. Briefly, the original splitting matrix $M$ has multiple rows that contain only $0$ or $1$. This causes the sparsity of the null-space vectors and the computed $ZZ^T$. This is detectable by the AP so AP must refuse to compute $W$. 
       {\remark \label{rem:explainY} To handle this, Algorithm \ref{alg:Secure-Kernel} adds columns with random numerical values to $M$ before generating the null-space vectors.
      \[M^* = [M ~M' ~Y] \in \{0,1\}^{n \times (l + l_1 + l_2)}\]}
        where $Y \sim \mathcal{N}(\mu, \sigma) \in \mathbb{R}^{n \times l_2}$. 
        Then AP applies Algorithm \ref{alg:Secure-Response} to generate a secure response $W$.
       Note that the columns of $M^*$ are independent so the multiplication result of columns in $M'$ and $Y$ are omitted in the final evaluation.
       
    \end{enumerate}

    From the analysis, we conclude that FedXGBoost-SMM-2 satisfies the privacy-preserving constraints. Furthermore, it is more secure than SecureBoost and FedXGBoost-SMM-1 under the assumption that the curious party has unbounded computational capability.  

    \subsection{FedXGBoost-LDP}
    FedXGBoost-LDP is a different approach from FedXGBoost-SMM  that perturbs the gradients and hessians to achieve privacy-preserving, e.g from \cite{Duchi}, Piecewise or Hybrid mechanisms from \cite{Wang}, etc. The perturbed data is then used directly for training, which reduces the training time in comparison to HE or SMM methods. Nevertheless, due to the high nonlinearity of the splitting score function, the injected noise degrades the utility strongly. For this reason, we use the first-order approximation for (\ref{eqn:optimizeFunc}) to evaluate the splitting score. Appendix \ref{appendix:firstOrder} depicts that the first-order approximation offers an unbiased estimator for the splitting score, and the optimal splitting candidate from the set $\mathcal{S} = \{s_1, ~\cdots, ~s_l\}$ is determined as
    \begin{equation} \label{eqn: Lsplit_short_first_order}
    \begin{split}
        s^* &= \arg \max_{s_i \in \mathcal{S}} -\frac{1}{\lambda}G^L_{s_i} G^R_{s_i}
    \end{split}
    \end{equation}
    Appendix \ref{appendix:firstOrder} also depicts that the variance of the evaluation depends strongly on the variance of the injected noise, which implies a compromise between privacy and the accuracy. Nevertheless, this method accelerate the training process significantly in comparison to encryption or linear algebra techniques, thus it is evaluated as an heuristic approach. Procedure \ref{alg:LDPFedXGBoost} describes how the best splitting score is estimated by FedXGBoost-LDP. The procedure is similar to the plain XGBoost, the only difference is the regression training uses $g^*, h^*$ and evaluates the splitting candidates by (\ref{eqn: Lsplit_short_first_order}).
    
\begin{algorithm} [tb] 
\caption{FedXGBoost-LDP}
\label{alg:LDPFedXGBoost}
\textbf{Input:}
\begin{itemize}
    \item Users set $\mathcal{N}_I$, with $|\mathcal{N}_I| = n$ 
    \item Active party (\textbf{AP}) has $g, ~h \in \mathbb{R}^n$
    \item Total $p$ passive parties ($\textbf{PP}$), each has $k^{th}$ PP has data base $D^k \in \mathbb{R}^{n \times d}$ and constructs a private set of $l$ splitting candidates $\mathcal{S}^k = \{s_1 \cdots s_l\}$
    \item $\mathcal{M}(.): \mathbb{R}^n \rightarrow \mathbb{R}^n$: LDP perturbation mechanism
\end{itemize}
\textbf{Output of protocols:} $\operatorname*{argmax}_{s_i \in \mathcal{S}^k, ~k \in \{1 \cdots p\}} L^{s_i}_{split}$ \\
\textbf{Procedures:}
\begin{algorithmic}[1] 
\STATE \textbf{AP:} $\{g^*\}\xleftarrow{} \mathcal{M}(g), ~G  \leftarrow \sum_{i \in \mathcal{N}_I}g_i$   
\STATE \textbf{AP:} Transfer $\{g^*,~G\}$ to PP.
\FOR{k = 1 \textbf{to} p} 
\STATE $\textbf{PP}_{k}:$
\STATE $G^L = \{G^L_{s_i}\}_{i \in \{1, \cdots, l\}} \leftarrow M^Tg^*$ 
\STATE $(L^k)^* \leftarrow -\infty, ~(s^k)^* \leftarrow 0$
\FOR{$s_i = s_1 ~\textbf{to} ~s_l$} 
\STATE $G^R_{s_i} = G - G^L_{s_i}$ 
\STATE $L \leftarrow  -\frac{1}{\lambda}G^L_{s_i} G^R_{s_i}$
\IF{$(L^k)^* < L$}
    \STATE $(L^k)^* \leftarrow L$
    \STATE $(s^k)^* \leftarrow s_i$
    \ENDIF
\ENDFOR
    \STATE Transfer $(L^k)^*$ to AP
\ENDFOR 
\STATEx /* Evaluate optimal splitting score over all parties */
\STATE \textbf{AP:} 
\STATEx Find $k^* = \operatorname*{argmax}_{k \in \{1 \cdots p\}} L^k$
\STATEx Request $s^* = (s^k)^*$ from PP$_{k^*}$
\end{algorithmic}
\textbf{Return:} $\text{PP}_{k^*}, ~(s^k)^*$
    
\end{algorithm}
    
\section{The complete FedXGBoost Protocols - FedXGBoost-SMM \& FedXGBoost-LDP} \label{sec:Prot}
Firstly, the AP determines the set of users (the users in a common node) being analyzed and announces this to all PPs. Next, all parties follow either FedXGBoost-SMM or FedXGBoost-LDP to find the best splitting candidate.
After determining optimal splitting score, AP requests the information from the owner of the best score. Particularly, AP requests the feature analyzed by that splitting operation and the set of users in left and right nodes. 
After receiving the feature information, AP constructs a look-up table to record the corresponding PP and the analyzed feature. On the other side, the chosen PP also records the chosen feature and the best splitting operation for the usage in the regression phase. The training process continues from the split users' space until the tree reaches the maximum depth. At this stage, optimal leaf weight is computed according to (\ref{eqn:optimalWeight}) and saved for the prediction phase. This completes the construction of one regression tree. 

The application of the regression of the trained model is similar to the study by \cite{Cheng}. When the AP wants to make inference from a new instance, it uses its look-up table and cooperates with the PPs to determine in which tree leaf the instance belongs to. Next, it aggregates the optimal weight overall regression trees to obtain a final prediction. 

\section{Experiments and Evaluation} \label{sec:Exp}
We provide the experiments on two dataset 1) "Give Me Some Credit" \footnote{https://www.kaggle.com/c/GiveMeSomeCredit} and 2) "Default of Credit Card Clients" \footnote{https://www.kaggle.com/uciml/default-of-credit-card-clients-dataset} published on Kaggle. The first dataset contains the data of 150000 users with 10 attributes. The second dataset contains the data of 30000 users with 25 attributes. The experiments are conducted by two 16-cpu-core 64GB-memory Linux machines with network bandwidth of 25000Mb/s.  Leverage Apache Spark platform is applied for distributed computing. We evaluate FedXGBoost-LDP by comparing the model accuracy and the time consumption with the plain XGBoost and the studied encryption techniques. FedXGBoost applies different LDP perturbation techniques, which are Laplace mechanism (LM) from \cite{Dwork} and Duchi's method (DM) from \cite{Duchi} with varying privacy budget $\epsilon = \{1,3\}$. According to (\ref{eqn: Lsplit_short_first_order}), we also evaluate LDP mechanisms using first-order approximation, which is expected to reduce the accuracy loss by the injected noise. 

\begin{figure} [tb]
\centering
\includegraphics[width=80mm]{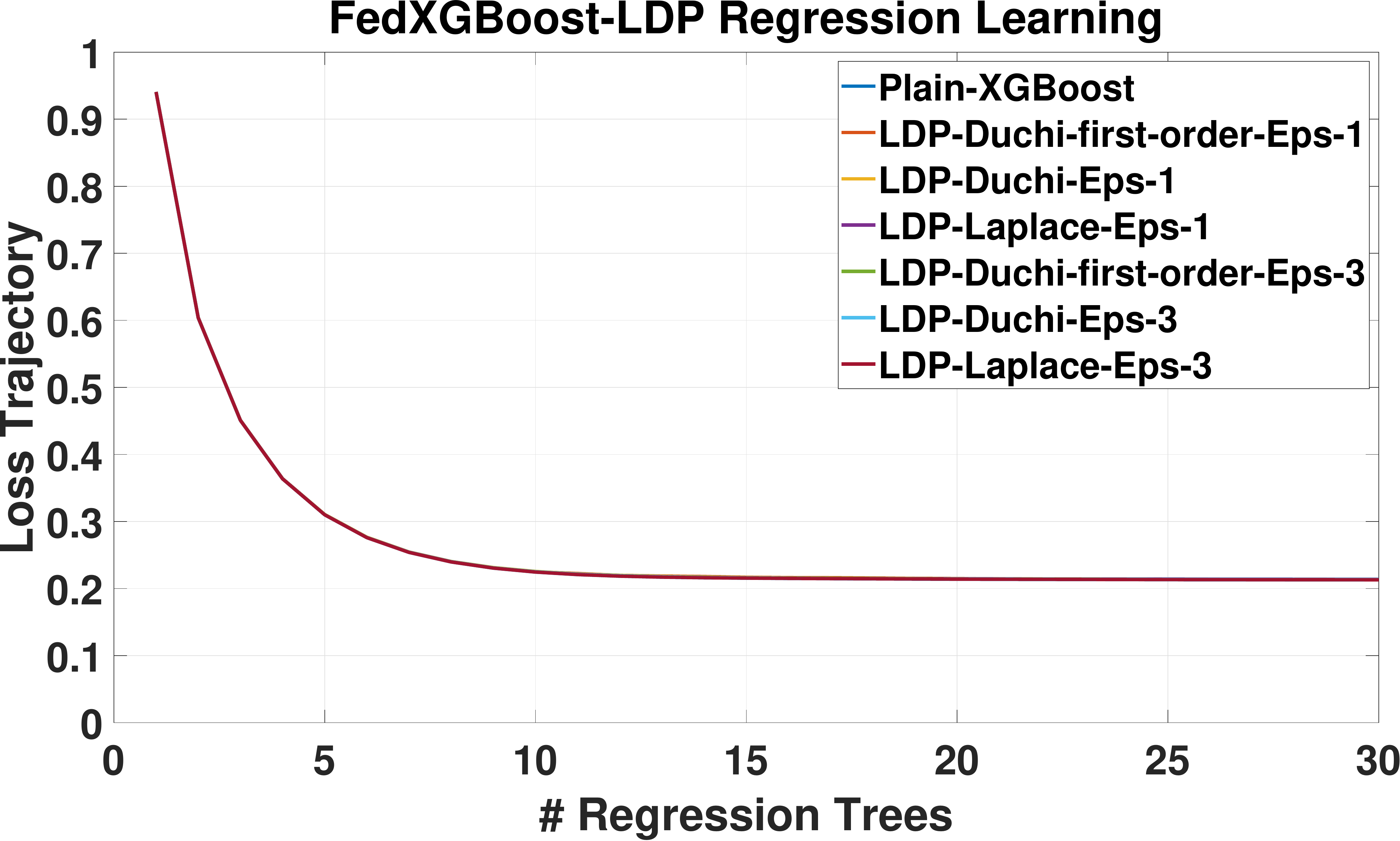}
\caption{Loss trajectory of FedXGBoost-LDP learning phase on ``Give Me Some Credit" dataset with 150000 instances}
\label{fig:LDP150k}
\end{figure}
\begin{figure} [H]
\begin{subfigure}{.5\textwidth}
\centering
\includegraphics[width=80mm]{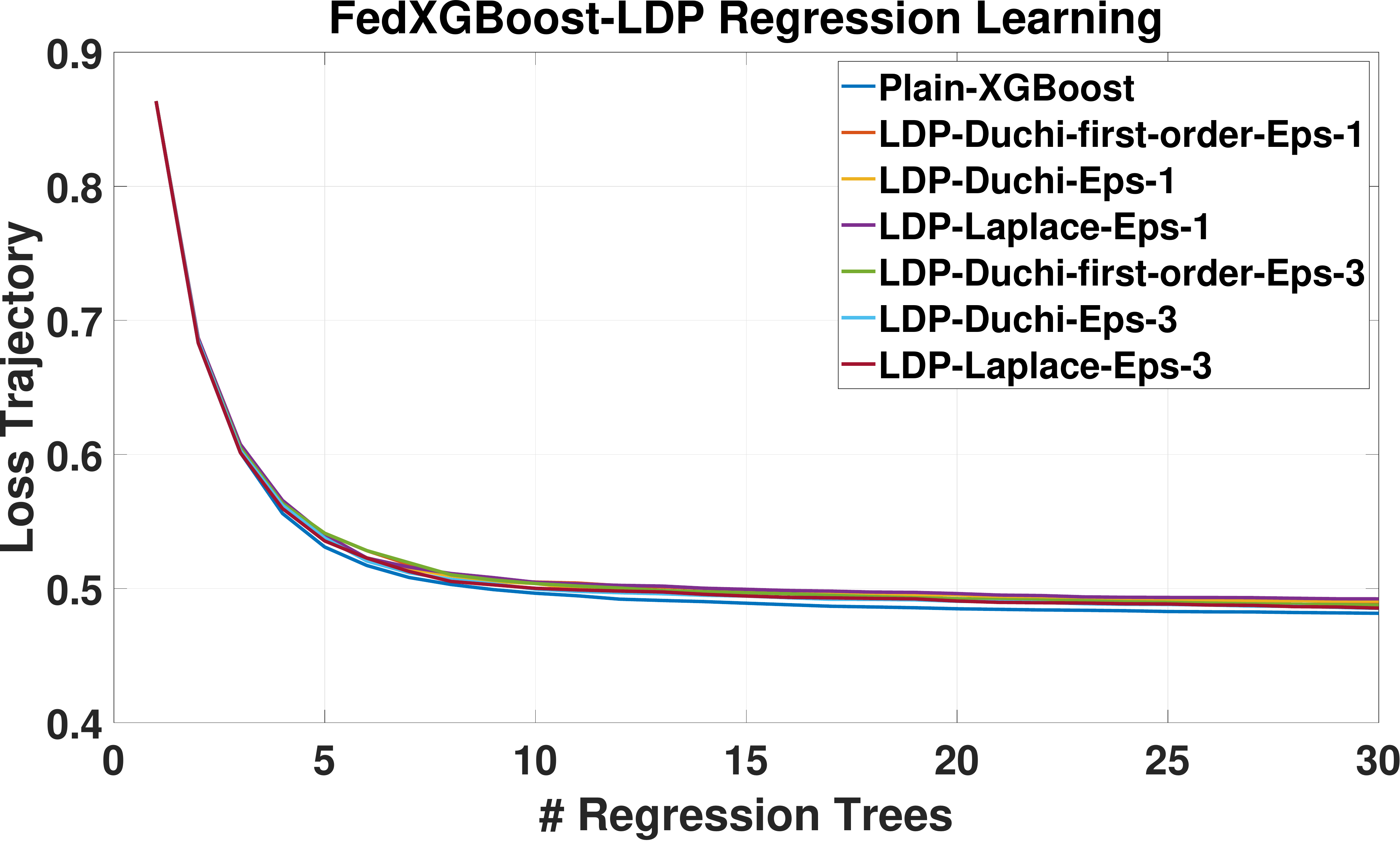}
\caption{Overall learning trajectory}
\end{subfigure}
\begin{subfigure}{.5\textwidth}
\centering
\includegraphics[width=80mm]{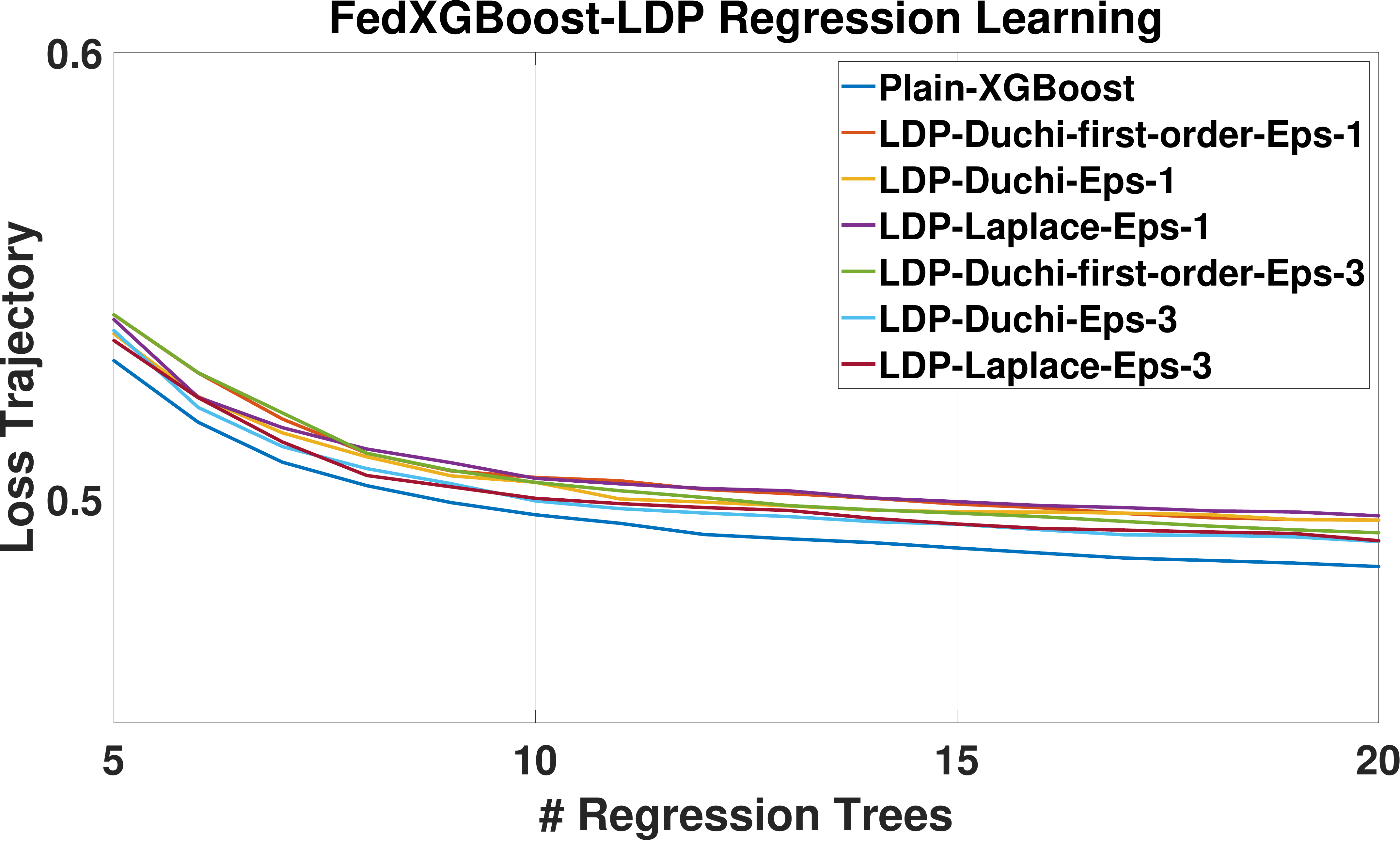}
\caption{Convergence rate analysis: from the $5^{th}$ to the $20^{th}$ iteration}
\label{fig:LDP30kFocus}
\end{subfigure}
\caption{Loss trajectory of FedXGBoost-LDP learning phase on ``Default of Credit Card Clients" dataset with 30000 instances.}
\label{fig:LDP30k}
\end{figure}

Figure \ref{fig:LDP150k} and \ref{fig:LDP30k} depict the loss trajectory of two experiments. As can be seen from figure \ref{fig:LDP30kFocus}, despite the injected noise cause a small performance reduction, the loss convergence is equivalent to the plain XGBoost. Table 1 and 2 evaluate the time consumption of different approaches. In general, LDP methods have negligible overhead in comparison with plain XGBoost. Compared with encryption method, it tremendously accelerate the training process.

\begin{table}[tb]
\label{tab:eval150k}
\begin{center}
\begin{tabular}[H]{ |c|c|c|c| } 
\hline
Method & Parameter & \specialcell{Time [s] \\ n = 30K } & \specialcell{Time [s] \\ n = 150K } \\
\hline
XGBoost & - & 69 & 260  \\ 
\hline
Paillier Encryption & - & 354 & 1560 \\
\hline
\specialcell{FedXGBoost-LDP} & LM: $\epsilon = 1$ & 82 & 270\\ 
\hline
\specialcell{FedXGBoost-LDP} & \specialcell{DM: $\epsilon = 1$} & 71 & 240\\ 
\hline
\specialcell{FedXGBoost-LDP} & \specialcell{DM: $\epsilon = 1$ \\ First Order}  & 67 & 212\\ 
\hline
\specialcell{FedXGBoost-LDP} & LM: $\epsilon = 3$ & 76 & 270\\ 
\hline
\specialcell{FedXGBoost-LDP} & \specialcell{DM: $\epsilon = 3$} &  73 & 235\\ 
\hline
\specialcell{FedXGBoost-LDP} & \specialcell{DM: $\epsilon = 3$ \\ First Order} & 70 & 204\\ 
\hline
\end{tabular}
\caption{Training time consumption for one regression tree of different approaches}
\end{center}
\end{table}

\section{Conclusion} \label{sec:Con}
This paper studies two different protocols (FedXGBoost-SMM and FedXGBoost-LDP) that enable the state of the art tree ensemble model XGBoost to be conducted under FL settings. Different from the previous work applying homomorphic encryption, our linear algebra based protocol FedXGBoost-SMM incurs lower overhead while maintaining lossless accuracy. We also propose and empirically evaluate the accuracy of the heuristic protocol FedXGBoost-LDP, which relaxes the splitting score computation to first order approximation for computational speedup, and uses  LDP noise perturbation. For  future work, we will experimentally evaluate the overhead of FedXGBoost-SMM. Further study of scalable privacy-preserving XGBoost for FL is crucial for its deployment in practice. 

\clearpage
\appendix
{\Large\textbf{Appendix}}

\section{Unbiased splitting score estimation by using first-order approximation for the regression objective} \label{appendix:firstOrder}
Due to the high nonlinearity of (\ref{eqn:optimizeFunc}), the noise injected gradients and hessians is expected to degrade the regression learning strongly. This appendix depicts that if we apply the first-order approximation for the regression objective, we can use an unbiased estimator to evaluate the noisy splitting score. 

\begin{equation} \notag
\begin{split}
     \mathcal{L}^{(t)} & = \sum_{i = 1}^n (l( y_i, \hat{y}^{(t-1)}_i + f_t(x_i)) + \Omega(f_t)\\
    & \approx \sum_{i = 1}^n (l(y_i, \hat{y}^{(t-1)}_i) + g_i f_t(x_i)) + \Omega(f_t)\\
    & = \mathcal{L}^{(t-1)} + \sum_{i = 1}^n (g_i f_t(x_i)) + \gamma T + \frac{1}{2}\lambda \sum_{j = 1}^T w_j^2\\
    & = \mathcal{L}^{(t-1)} + \sum_{j = 1}^T\left[(\sum_{i \in \mathcal{N}_j}g_i) f_t(x_i)\right] + \gamma T + \frac{1}{2}\lambda \sum_{j = 1}^T w_j^2\\
    & = \mathcal{L}^{(t-1)} + \sum_{j = 1}^T\left[(\sum_{i \in \mathcal{N}_j}g_i) w_j + \frac{1}{2}\lambda w_j^2\right] + \gamma T\\
\end{split}
\end{equation}
From here, we obtain
\begin{equation} \notag
\begin{split}
    \mathcal{L}^{(t)} = \mathcal{L}^{(t-1)} + \sum_{j = 1}^T\left[(\sum_{i \in \mathcal{N}_j}g_i) w_j + \frac{1}{2}\lambda w_j^2\right] + \gamma T\\
\end{split}
\end{equation}
In the above approximation, $f_t(x_i)$ is the prediction of the new constructed tree, which is the weight $w_j$ of the node $j$. The function achieves minimum when the following condition is fulfilled
\[w_j = -\frac{1}{\lambda}(\sum_{i \in \mathcal{N}_j}g_i)\]
The obtained objective from the optimal weight is
\[\mathcal{L}^{'} = - \frac{1}{2}\sum_{j = 1}^T \frac{(\sum_{i \in \mathcal{N}_j}g_i)^2}{\lambda} + \gamma T\]
Since the nodes' weight decide how well the prediction will be, the splitting score can be formulated as the winning in the prediction lost after splitting the users set $\mathcal{N}_j$ into two subsets $\mathcal{N}_L, ~\mathcal{N}_R$. The preicition gain by different splitting candidates $s_i$ are evaluated by the function $\mathcal{L'}_{split}$ as
\begin{equation} \notag
    \begin{split}
        \mathcal{L}^{'}_{split} & = \frac{1}{2 \lambda}  \left[(\sum_{i \in \mathcal{N}_L}g_i)^2 + (\sum_{i \in \mathcal{N}_R}g_i)^2 - (\sum_{i \in
        \mathcal{N}_j}g_i)^2 \right] - \gamma  \\
        & = \frac{1}{2 \lambda}  \left[(G^L_{s_i})^2 + (G^R_{s_i})^2 - G^2 \right] - \gamma \\
    \end{split}
\end{equation}
{\remark We are evaluating only the new tree so the aggregated results from the $T-1$ constructed trees are not considered.}
Finding the optimal splitting candidate is equivalent to
\begin{equation} \label{eqn: Lsplit_short_first_order_appen}
\begin{split}
    s^* &= \arg \max_{s_i \in \mathcal{S}} -\frac{1}{\lambda}G^L_{s_i} G^R_{s_i} \\
\end{split}
\end{equation}

In comparison to the splitting score function obtained from the second order approximation, the function (\ref{eqn: Lsplit_short_first_order_appen}) does not consider the hessians in the denominator, thus the estimated result for the noisy gradients is proven to be unbiased.
{\proof
Let the noise injected gradients for each instance $i$ be $X^g_i$ with the following properties:
\[E[X^g_i] = g_i\]
\[Var[X^g_i] = \sigma^2_i\]
The aggregated gradients of the left and right nodes by a splitting candidate $s_i$ is represented by
\[X_{G^L_{s_i}} = \sum_{i \in \mathcal{N}_L}X^g_i, ~X_{G^R_{s_i}} = \sum_{i \in \mathcal{N}_R}X^g_i\]
 To simplify the analysis, we omit the index term $s_i$ of each splitting candidate. These estimator for the aggregated gradients have the following statistical properties
 \[E\left[X_{G^L}\right] = E\left[\sum_{i \in \mathcal{N}_L}X^g_i\right] = \sum_{i \in \mathcal{N}_L}E\left[X^g_i\right] = \sum_{i \in \mathcal{N}_L}g_i = G^L,\]
 \[E\left[X_{G^R}\right] = E\left[\sum_{i \in \mathcal{N}_R}X^g_i\right] = \sum_{i \in \mathcal{N}_R}E\left[X^g_i\right] = \sum_{i \in \mathcal{N}_R}g_i = G^R,\] 
\[Var[X_{G^L}] = \sum_{i \in \mathcal{N}_L} \sigma^2_i, ~Var[X_{G^R}] = \sum_{i \in \mathcal{N}_R} \sigma^2_i\]
The goal is to estimate the splitting score according to (\ref{eqn: Lsplit_short_first_order_appen}) by the estimator $X_{\mathcal{L}^{'}}$ as
\begin{equation} \label{eqn:firstEst}
{X_{\mathcal{L}^{'}}} =  -\frac{1}{\lambda} {X_{G^L}} {X_{G^R}} 
\end{equation}
The estimator is unbiased as can be shown by \\
\[E[X_{\mathcal{L}^{'}}] = -\frac{1}{\lambda} E[X_{G^L}] E[X_{G^R}] = -\frac{1}{\lambda} G_L G_R\]
Its variance is
\begin{equation} \notag
    \begin{split}
      Var[X_{\mathcal{L}^{'}}] & = \frac{1}{\lambda^2}\left(E[X_{G^L}]^2Var[X_{G^R}] + E[X_{G^R}]^2Var[X_{G^L}] \right) \\
      & = \frac{1}{\lambda^2}\left((G^L)^2\sum_{i \in \mathcal{N}_L} \sigma^2_i + (G^R)^2\sum_{i \in \mathcal{N}_R} \sigma^2_i\right)
    \end{split}
\end{equation}

It is shown that the estimator from (\ref{eqn:firstEst}) is unbiased but has large variance. 
}
\clearpage
\bibliographystyle{named}
\bibliography{ijcai21}

\end{document}